\documentclass[singlecolumn]{article}
\usepackage{amssymb}
\usepackage{multirow,setspace,verbatim,amsfonts,graphicx,amsmath,amsthm,amsbsy,amssymb,epsfig,url}
\usepackage{amsthm}
\usepackage{gensymb}
\usepackage{booktabs}
\usepackage{cite}
\usepackage{makecell}
\usepackage{epstopdf}

\usepackage{algorithm}
\usepackage[noend]{algpseudocode}
\usepackage{amsfonts}
\usepackage{amsmath,bm}
\usepackage{amssymb}
\usepackage{amsthm}
\usepackage{tikz,graphicx}
\usepackage{mathrsfs} 
\usepackage[algo2e]{algorithm2e} 

\usepackage{array,multirow}
\usepackage{color}
\usepackage{caption}
\usepackage{subcaption}

\newcommand{\fb}{{\mathbf f}}
\newcommand{\gb}{{\mathbf g}}

\newcommand{\Fc}{{\mathcal F}}

\theoremstyle{definition} \theoremstyle{plain}

\begin{document}
\begin{titlepage}
\begin{center}
\begin{LARGE}
\begin{bf}

Deep  artifact learning for compressed sensing and parallel MRI 

\end{bf}
\end{LARGE}
\end{center}

\smallskip

\begin{center}
\begin{large}
{\sc  Dongwook Lee$^{1}$,  Jaejun Yoo$^1$ and Jong Chul Ye$^{1,*}$}
\\[0.1in]
\end{large}

\smallskip

{\em $^1$Dept. of Bio and Brain Engineering\\
Korea Advanced Institute of Science \& Technology  (KAIST) \\\
373-1 Guseong-dong  Yuseong-gu, Daejon 305-701, Republic of Korea \\

Email: jong.ye@kaist.ac.kr \\
  \vspace*{0.2cm}
}
\end{center}

\noindent{
Running Head: Deep aliasing artifact learning for CS and parallel MRI \\
Journal: Magnetic Resonance in Medicine\\}

\noindent{
$\star$ Correspondence to:\\
Jong Chul Ye,  Ph.D. ~~\\
Professor \\\
Dept. of Bio and Brain Engineering,  KAIST \\
373-1 Guseong-dong  Yuseong-gu, Daejon 305-701, Korea \\
Email: jong.ye@kaist.ac.kr \\
Tel: 82-42-350-4320 \\
Fax: 82-42-350-4310 \\
\\
\\
Total word count : approximately 5000 words.

}

\end{titlepage}

\begin{abstract}

\noindent\textbf{Purpose:} 
Compressed sensing MRI (CS-MRI) from single and parallel coils is one of the powerful ways to reduce the scan time of MR imaging with performance guarantee. However, the computational costs  are usually expensive. This paper aims to propose a computationally fast and accurate deep learning algorithm for the reconstruction of MR images from highly down-sampled k-space data.\\
 \textbf{Theory:}
Based on the topological analysis, we show that
 the data manifold of the  aliasing artifact is  easier to learn from a uniform subsampling  pattern with additional low-frequency k-space data.
Thus,  we develop deep aliasing artifact learning networks for the magnitude and phase images to estimate and remove the aliasing artifacts from highly accelerated MR acquisition. \\
\textbf{Methods:} 
The aliasing artifacts are directly estimated from
the distorted magnitude and phase images  reconstructed from subsampled k-space data so that
we can get an aliasing-free  images by subtracting the estimated aliasing artifact from corrupted inputs. 
 Moreover, to deal with the globally distributed aliasing artifact, we  develop a multi-scale deep neural network  with a  large receptive field. 
 \\
\textbf{Results:} The experimental results confirm that the proposed deep artifact learning  network effectively estimates and removes the aliasing artifacts. Compared to existing  CS methods from single and multi-coli data, the proposed network shows minimal errors by removing the coherent aliasing artifacts. Furthermore, the computational time is by order of magnitude faster.
\\
\textbf{Conclusion:} 
As the proposed deep  artifact learning network immediately generates accurate reconstruction,   it has great potential for clinical applications.
\\

\vspace{15 mm}

\end{abstract}
\pagebreak

\newpage

\section*{Introduction}
MR imaging is one of the most valuable imaging methods in the clinic for the needs of diagnostic and therapeutic indications. However, the physical and physiological constraints basically limit the rate of MR acquisition. Since 
the long scan time is one of the shortcomings of  MR imaging, the efficient acceleration scheme for MR acquisition is important to reduce the acquisition time.
Accordingly, under-sampling of k-space is necessary, and many researchers have developed various reconstruction methods such as parallel imaging~\cite{pruessmann1999sense, griswold2002generalized} and compressed sensing MRI (CS-MRI)~\cite{donoho2006compressed,lustig2008compressed} that allow for accurate reconstruction from the insufficient k-space samples.

For example, generalized autocalibrating partial parallel acquisition (GRAPPA)\cite{griswold2002generalized} is a representative parallel MRI (pMRI) technique that interpolates the missing k-space data by exploiting the diversity of the coil sensitivity maps.
On the other hand, CS-MRI reconstructs a high-resolution image from randomly sub-sampled k-space data by utilizing the sparsity of the data in the transformed domain. CS algorithms are commonly formulated as penalized inverse problems that minimize the tradeoff between the data fidelity term in the k-space and the sparsity penalty in the transform domain.
The state-of-the-art CS algorithm in this field is the annihilating filter-based low-rank Hankel matrix approach (ALOHA), 
in which the CS-MRI and parallel MRI can be unified as an interpolation problem in the weighted k-space domain using a low-rank structured matrix completion~\cite{jin2016general,lee2016acceleration,lee2016reference,ye2016compressive}.
One of the limitations of these algorithms, however, is the computational complexity. Although these CS algorithms can achieve state-of-the-art performance, computational complexity is usually high and leads to an increase in the time for image reconstruction.
Moreover, the incoherent scanning patterns required for the CS-MRI are usually different from those of the standard acquisition, so additional pulse sequence programming is often required.

Recently, deep learning has proved to be an important framework for computer vision research \cite{krizhevsky2012imagenet}, thanks to the massive datasets and the development of computing hardwares such as GPUs.
In particular, convolution neural network ({\color{black}CNN}), which is a particular form of feedforward neural networks, use trainable filters (or weights) and convolution operations between the input and filters. CNN implicitly learns the filter coefficients to effectively extract local features from the training data.
Thus, CNN has a strong ability to capture the local features of the input images, resulting in a great success in classification problems~\cite{krizhevsky2012imagenet},  as well as regression problems such as segmentation~\cite{ronneberger2015u}, denoising~\cite{mao2016image, zhang2016beyond}, super resolution~\cite{dong2014learning}, etc. 

The success of deep learning has been recently investigated in statistical learning literatures \cite{vapnik1998statistical}, showing  that the exponential expressivity or representation power has been attributed to its success \cite{telgarsky2016benefits, bianchini2014complexity}.
Therefore, this paper aims to utilize the great ability of CNN to capture the feature of image structures for medical image reconstruction problems.

 In X-ray computed tomography (CT),
there are some successes in the applications of deep learning in image reconstruction. Kang et al.~\cite{kang2016deep} showed that the deep convolutional neural network (CNN) can efficiently remove the noises originated from the low-dose CT. They used the directional wavelets with deep CNN structure to reduce these low-dose artifacts. On the other hand, there were some studies in the sparse views  CT to remove the globally distributed streaking artifacts originating from the limited number of projection views~\cite{jin2016deep, han2016deep}.

In MRI, the first try to apply the deep learning approach was carried out by Wang et al~\cite{wang2016accelerating}. They have trained the deep neural network from the downsampled reconstruction images to learn a fully sampled reconstruction. They then attempted to combine the deep learning outcome with CS-MRI reconstruction methods in two ways. First, they used the image generated by the learned network to initialize the CS-MRI and then reconstructed images. Secondly, they used the output of the network as a reference image and used it as an additional regularization term in classical CS approaches. 
 Hammerinik et al.~\cite{hammernik2016learning} developed a deep network architecture as unfolded  iterative steps of CS-MRI.  Instead of using the classical regulizers such as $l_1$ and total varitaion, their networks learned a set of filters and corresponding penalty functions using a reaction diffusion model~\cite{chen2015learning, chen2015trainable}. All the parameters of the network including the filters and the influence functions are trained from the set of training data.

\begin{figure}[!hb] 	
\center{\includegraphics[width=16 cm]{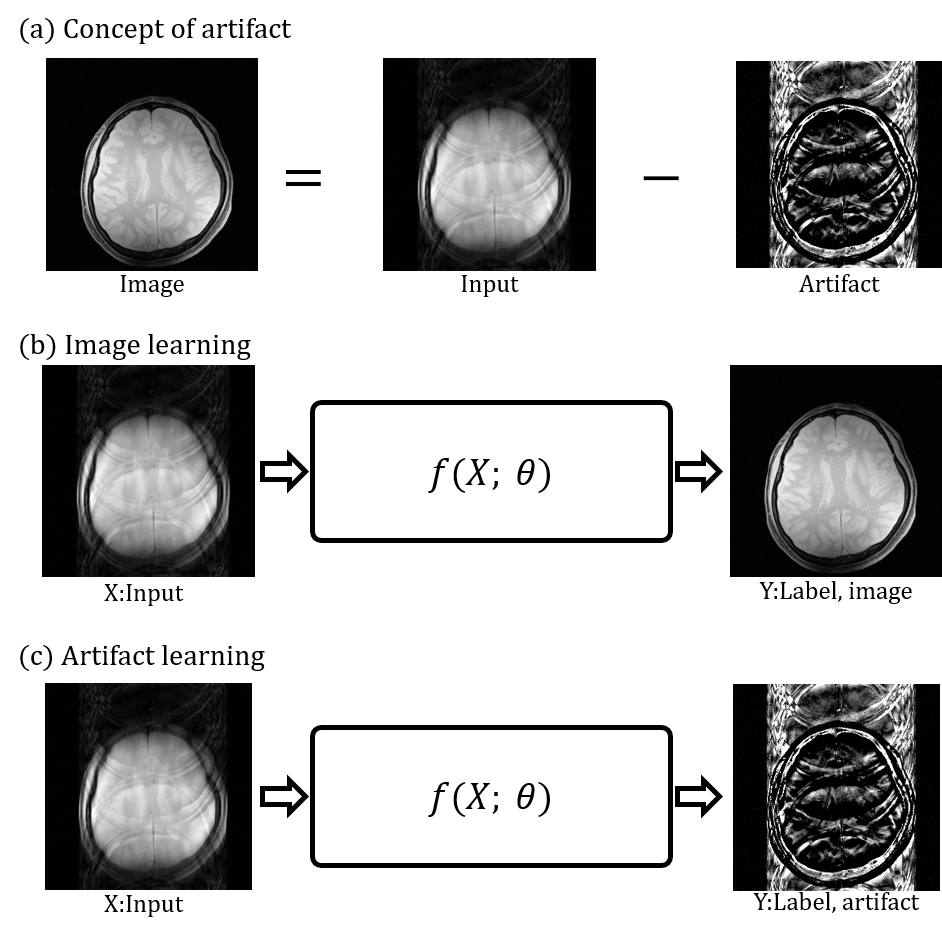}}
\caption{ Concept of artifact learning. (a) The artifact image is defined as the difference between the aliased image and the artifact-free image in magnitude and phase domain. (b) Image learning: the aliased image is mapped to the artifact-free images. (c) Artifact learning: the aliased image is mapped to the artifact image. Once the artifact image is estimated, the artifact corrected image can be obtained by subtracting the estimated artifact from the input image. 
}
\label{fig:concept}
\end{figure}

In his seminal work on statistical learning theory \cite{vapnik1998statistical}, Vapnik showed that the learning problem is a highly ill-posed inverse problem of the unknown  {\em probability distribution.}
Inspired by this insight and the similarity with the compressed sensing problem,
we are interested in adopting learning theory to expand the theory of compressed sensing to an inverse problem of the unknown {\em  distribution}, rather than  the inverse problem of  a single realization.
This view gives us much flexibility and clear directions in designing learning architecture for a given compressed sensing MR problem.
Specifically, to allow for an accurate distribution estimate, similar to the role of sparse transformation in CS theory, we must find a way to simplify the data distribution to meet the representation power of a given network.
Among various possible methods, this paper examines various sub-sampling patterns, the associated aliasing artifact patterns, and the image against the artifact data manifold to find out what makes the data distribution simple.
Specifically, using a computational topology  called persistent homology, we show that the aliasing artifact from uniform subsampled k-space data with a few low-frequency components
has a  simpler topological structure, so that learning these artifacts 
is  easier than learning the original artifact-free images.
This causes us 
 to learn aliasing artifacts  rather than the aliasing-free image from fully sampled data (see Fig.~\ref{fig:concept} (b), (c)).
Once the aliasing artifacts are estimated, an aliasing-free image is then obtained by subtracting the estimated aliasing artifact as shown in Fig.~\ref{fig:concept} (a). 

Another important contribution of this paper is that for the globally distributed aliasing artifacts{\color{black},} it is shown that a desired neural network should have a large receptive field to cover the entire artifacts. As a possible realization of the large receptive field,   the deconvolution network \cite{noh2015learning} 
 with contracting path, often referred to as a  
U-net structure~\cite{ronneberger2015u}  \cite{he2015deep},  is shown  effective in estimating the aliasing artifacts. 
By combining the artifact learning scheme with the  U-net structure,  we  show that the reconstruction performance can be significantly improved. 

Furthermore,  another very important advantage of the proposed method is that, once the network is trained,
 the network immediately produces accurate results while the existing CS algorithms require substantially higher time and computing costs.
One might argue that the training time should be counted  as the reconstruction time  for a fair comparison. However, as long as the training can be carried out once with extensive {\color{black}datasets} (eg. on the manufacturer side), the reconstruction can be carried out immediately for each scanner. This is how current deep learning-based image or speech recognition systems were developed by Google, Baidu, etc., and the success of the proposed method informs us that this type of approach might be possible in the MRI problems.

\section*{Theory}

\subsection*{Problem Formulation}

In a multi-channel CS-MRI problem, 
the k-space measurement data is given by
\begin{eqnarray}\label{eq:fwd}
G= A F  
\end{eqnarray}
with
$$G=\begin{bmatrix}\gb_1 & \gb_2 & \cdots &\gb_C\end{bmatrix}, ~F=\begin{bmatrix}\fb_1 & \fb_2 & \cdots & \fb_C\end{bmatrix}$$
where $\fb_i$ and $\gb_i$ denote the unknown image and the corresponding k-space measurements from the $i$-th coil, respectively; $C$ is the number of coils; $A$ is a  subsampled Fourier matrix.
The minimum norm reconstruction from sub-sampled measurement can be obtained by
\begin{eqnarray}\label{eq:mn}
\hat F = A^\dag G, 
\end{eqnarray}
where $A^\dag=A^H(AA^H)^{-1}$ denoting the pseudo-inverse.  For  the cartesian trajectory, this can be easily obtained by taking the fast Fourier transform (FFT) after zero-padding k-space data.
However, the main issue is that the minimum norm solution suffers from aliasing artifacts.

To address this, the popular approaches is using compressed sensing approach by imposing the sparsity in the transform domain.
For example, $l_1$-SPIRiT ($l_1$- iTerative Self-consistent Parallel Imaging Reconstruction)  \cite{lustig2010spirit} utilizes the GRAPPA type constraint as an additional constraint  for a compressed sensing problem.
A recent ALOHA algorithm fully {\color{black}extends}  the insight of GRAPPA and converts CS-MRI and pMRI problem into a k-space interpolation problem using a low-rank interpolation using a structured matrix completion~\cite{jin2016general,lee2016acceleration,lee2016reference,ye2016compressive}.
However, to solve the CS problem, computationally expensive  iterative reconstruction methods must be  used.
Thus,  instead of solving the computationally expensive optimization problems,  this paper is mainly interested in obtaining the original image
$\{\fb_i\}_{i=1}^C$ from the aliased images \eqref{eq:mn} using machine learning approaches.

One possible approach is to learn the artifact free images from the aliased images (Fig.~\ref{fig:concept} (b)).
Specifically, suppose we are given a  sequence of  training data
$$S=\{(X_1, Y_1),\cdots, (X_n, Y_n)\}$$
where
$$(X_i, Y_i) = (\hat F^{(i)}, F^{(i)})$$
where $\hat F^{(i)}$ and $F^{(i)}$  denote the aliased image  defined in \eqref{eq:mn} and the artifact-free image in \eqref{eq:fwd}
for the $i$-the data, respectively. 
Then,  a neural network $f: X\rightarrow Y$ is trained such that it minimizes the empirical risk:
$$\hat L_n(f) = \frac{1}{n}\sum_{i=1}^n \|Y_i - f(X_i)\|^2$$

Another approach is to learn the aliasing artifact as the difference between aliased MR image and artifact-free image  (Fig.~\ref{fig:concept} (c)).
In particular, we define the artifact in the magnitude  and the phase domains separately, since it is easier to learn the real valued data than the complex-valued ones.
Specifically, the  magnitude  and phase domain artifacts are defined as follows:
\begin{eqnarray}
\begin{cases}   \quad R_{mag} = |\hat F| - |F| \\ 
 \quad R_{phase} = \angle \hat F - \angle F
\end{cases}
\label{eq:artifact}
\end{eqnarray} 
where $|\cdot| $ and $ \angle$ represent {\color{black}the element-wise} absolute and angle of a complex number.
Then, the training data is given by
$$S=\{(X_1, Y_1),\cdots, (X_n, Y_n)\}$$
where 
\begin{eqnarray*}
(X_i,Y_i) = \begin{cases}   (|\hat F^{(i)}, R_{mag}^{(i)})  &:  \text{(magnitude network)} \\ 
 (\angle \hat F^{(i)}, R_{phase}^{(i)})  &:  \text{(phase network)} 
  \end{cases}
\end{eqnarray*} 
for the $i$-the data. 
Again,  a neural network $f: X\rightarrow Y$ is trained such that it minimizes the empirical risk:
$$\hat L_n(f) = \frac{1}{n}\sum_{i=1}^n \|Y_i - f(X_i)\|^2$$

\subsection*{Generalization bound }

 However, the direct minimization of empirical risk, $\hat L_n(f)$, is problematic due to the potential  issue of the overfitting.
 In order to avoid overfitting,  we must minimize the risk
 $$L(f) = E_D \|Y-f(X)\|^2,$$ 
 where $E_D[\cdot]$ denotes the expectation under the data distribution $D$.
 However, the distribution $D$ is unknown, so we cannot directly minimiize the risk; instead, we are interested in bounding the risk with computable quantities.
 This is often called the {\em generalization bound} \cite{vapnik1998statistical}.

Specifically, 
the risk of a learning algorithm can be bounded in terms of complexity measures (eg. VC dimension and shatter coefficient) and the empirical
risk~\cite{vapnik1998statistical}. 
The Rademacher complexity~\cite{bartlett2002rademacher} is one of the most modern notions to measure the complexity that is distribution-dependent and defined for any class of  functions.
Specifically, with probability $\geq 1-\delta$,  
\begin{equation}\label{eq:L}
 L(f) \leq \underbrace{\hat L_n(f)}_{\text{empirical risk}} + \underbrace{2 \hat R_n(\Fc)}_{\text{complexity penalty}}+ 3 \sqrt{\frac{\ln(2/\delta)}{n}}
\end{equation}
where the empirical Rademacher complexity $\hat R_n(\Fc)$ is defined to be
$$\hat R_n(\Fc)=  E_\sigma \left[\sup_{f\in \Fc} \left(\frac{1}{n}\sum_{i=1}^n \sigma_i f(X_i) \right) \right],$$ 
where $\Fc$ denotes a functional space, and  $\sigma_1,\cdots, \sigma_n$ are independent random variable uniformly chosen from $\{-1,1\}$.
Therefore, in order to reduce the risk,   both the empirical risk (i.e.  data fidelity) and the complexity terms in Eq.~\eqref{eq:L} must be simultaneously minimized.

In a neural network, the value of  the risk is determined by the representation power of the network \cite{telgarsky2016benefits}, whereas the complexity term is determined by the structure of the network \cite{bartlett2002rademacher}. 
In fact,  the fundamental trade-off lies between the network complexity and empirical risk in the generalization bound.
When a deep network becomes more complex, the representation power increases to reduce the empirical risk at the expense of the increased complexity penalty.

So {\color{black}this} gives us a further perspective in the design of a deep network.
Specifically, if there is a means to convert the learning problem more easily, we can use a simpler network with less complexity penalty that can also reduce the empirical risk so that the generalization bound can be reduced.
In fact, it is shown that the complexity of a learning problem is determined by the complexity of the label data distribution.
This issue is discussed in detail in the following section.

\subsection*{Persistent homology analysis of data distribution}

Based on the above discussion of the fundamental trade-off, 
to measure the complexity of data distribution, which is a topological concept, we employ the recent computational topology tool called {\em persistent homology}~\cite{edelsbrunner2008persistent}.

\begin{figure}[!b] 	
\center{\includegraphics[width=17cm]{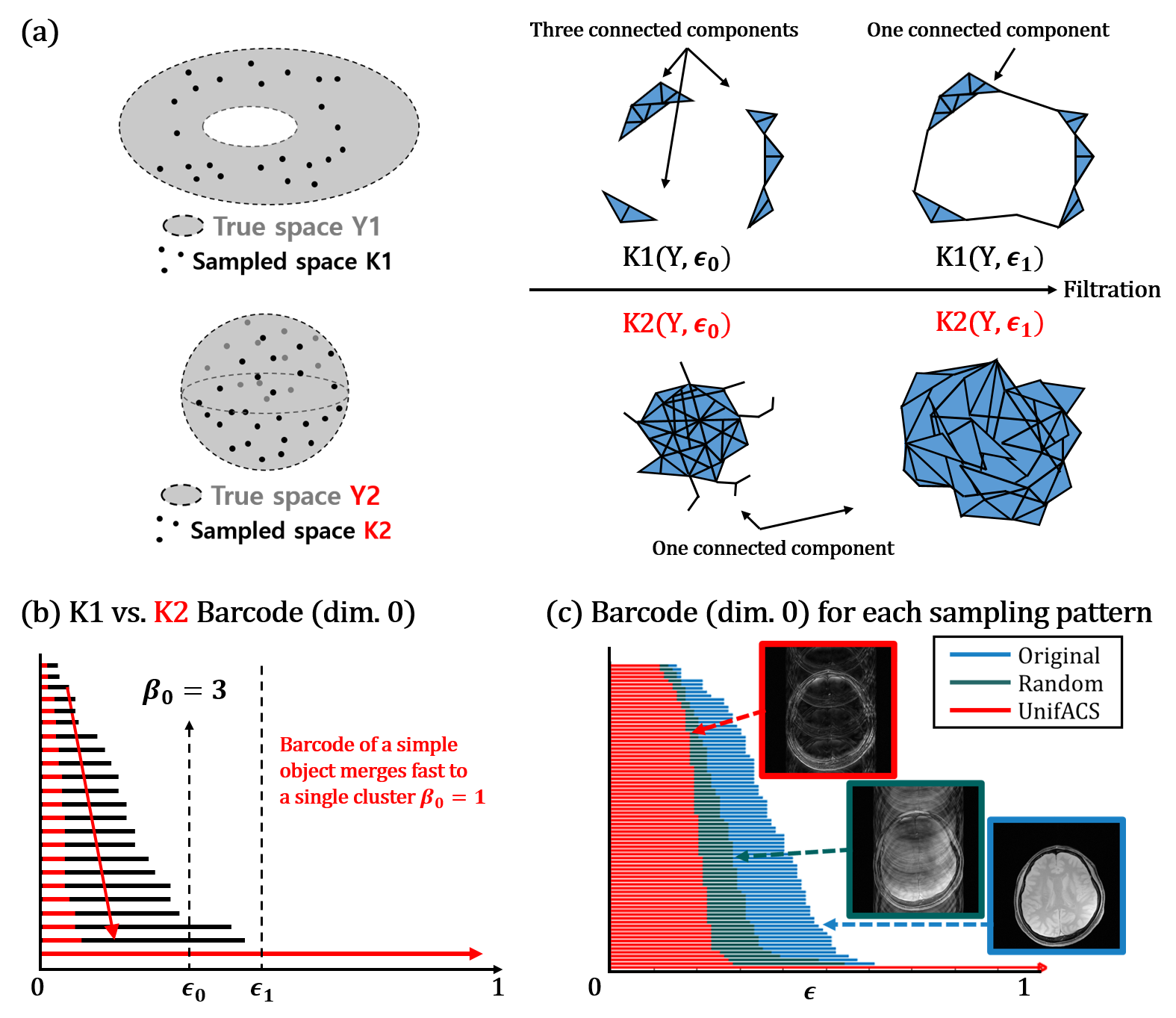}}
\caption{ (a) Point {\color{black}clouds} data $K1$ and $K2$  of the underlying data manifolds  $Y1$ and $Y2$. Their topological configurations with respect to the $\epsilon$-distance filtration show that the $K2$ merges faster to a single cluster than $K1$. (b) Zero-dimensional barcodes of the  point clouds data. The barcode of a simple object (i.e. $K1$) merged faster to a single cluster. (c) Zero-dimensional barcodes of the original image (blue), artifact image from Gaussian random down-sampling (green), and artifact image from uniform sampling with ACS lines (red) for the case of single-channel MR data at the acceleration factor of four. The barcode for the artifact image from uniform down-sampled k-space merges faster than the others. It means that the artifact data manifold from uniform sampling with ACS line has the simplest manifold.  
}
\label{fig:topology}
\end{figure}

In persistent homology, the topology of a space is inferred by investigating the change of multidimensional holes observed in different scales. Hole is an important topological characteristic which is invariant in the same topological class~\cite{edelsbrunner2008persistent}. Here, zero-dimensional hole is a connected components, one-dimensional hole is a cycle and two-dimensional hole is a void. For example,  in Fig.~\ref{fig:topology} (a), we can figure out that the topology of two spaces are different because the doughnut-like space ($Y1$) has a hole (i.e. a cycle) which the ball-like space ($Y2$) does not have. 

In practice, however, it is hard to infer the global topology of the continuous space directly with a discrete set of observed data (point clouds). As we differ the scale of observation by changing the distance measure $\epsilon$, the corresponding topology of the observed data space may vary (Fig.\ref{fig:topology} (a)). In persistent homology, instead of using a single fixed $\epsilon$, we investigate the space with entire $\epsilon$'s and find  holes which persist long over the evolutionary change. This process is called a  {\em filtration}. A hole which persists over varying scales is considered to be an important feature while the one with short persistence is considered as a topological noise. For example, a cycle which is an important feature of $Y1$ space will persist long while the three connected components at $\epsilon_0$ will disappear as allowable distance increases to $\epsilon_1$ (Fig. \ref{fig:topology} (a)). On the other hand, a connected component of the ball emerges early ($\epsilon=\epsilon_0$) and persists long till the end of filtration ($\epsilon=1$)~\cite{edelsbrunner2008persistent}.

During this filtration process \cite{edelsbrunner2008persistent}, the number of $m$-dimensional holes of a manifold called Betti numbers ($\beta_m$) are calculated. Specifically, $\beta_0$ represents the number of connected components.
As $Y1$ has more diverged and complicated topology than $Y2$, its point cloud merges slowly, which are reflected as a slow decrease in Betti numbers. This trend is illustrated using so-called {\em barcodes}~\cite{edelsbrunner2008persistent}.  As shown in Fig.\ref{fig:topology} (b), red barcodes which represent connected components of point clouds from $Y2$ topology quickly merges to a single cluster while the black barcodes still remain to be separated.

In {\color{black}the} experimental results section, we will confirm that the prediction from the topological analysis with Betti {\color{black}numbers} fully reflects reconstruction performance by neural networks.
More specifically,
we will  compare the topological complexity of the original and aliasing artifact image spaces by the change in the Betti numbers, as shown in Fig.~\ref{fig:topology} (c). The result clearly show that the manifold of the original images is topologically more complex than those of the artifact images with two different sub-sampling patterns (unifACS and random). Here, unifACS represents the uniform sampling with auto-calibration signal (ACS) on low frequency region. The points in artifact image manifold merge much earlier to a single cluster,  which informs that the underlying manifold has a  simpler topology than the original image space. The complexity of the artifact space is also varied by sampling patterns, as shown in Fig.\ref{fig:topology} (c). 
This persistent homology results are shown in accordance with the reconstruction performance, as will be shown later.

\vspace{-0.3cm}
\subsection*{Design of a deep network with better representation power}
\vspace{-0.2cm}
 

There have been several studies that explain the benefit of depth in neural networks \cite{telgarsky2016benefits, bianchini2014complexity}. In deep networks,  the representation power grows exponentially with respect to the number of layers, whereas it grows at most polynomially in shallow ones~\cite{telgarsky2016benefits}. 
Specifically, Telgarsky \cite{telgarsky2016benefits} derived the bound of representation power, which cannot be overcome by  shallow networks.
Therefore, with the same number of resources, theoretical results supports  a deep architecture being preferred to a shallow one, and it increases the performance of the network by reducing
the empirical risk in Eq.\eqref{eq:L}.

Another important components of representation power is determined by the {\color{black}receptive field size}.
The receptive field size is a crucial issue in many visual tasks, as the output must
respond to large enough areas in the image to capture information about large objects.
In the following, we describe our multi-scale artifact learning network with a large receptive field.

\begin{figure}[!b] 	
\centering
\includegraphics[width=16 cm]{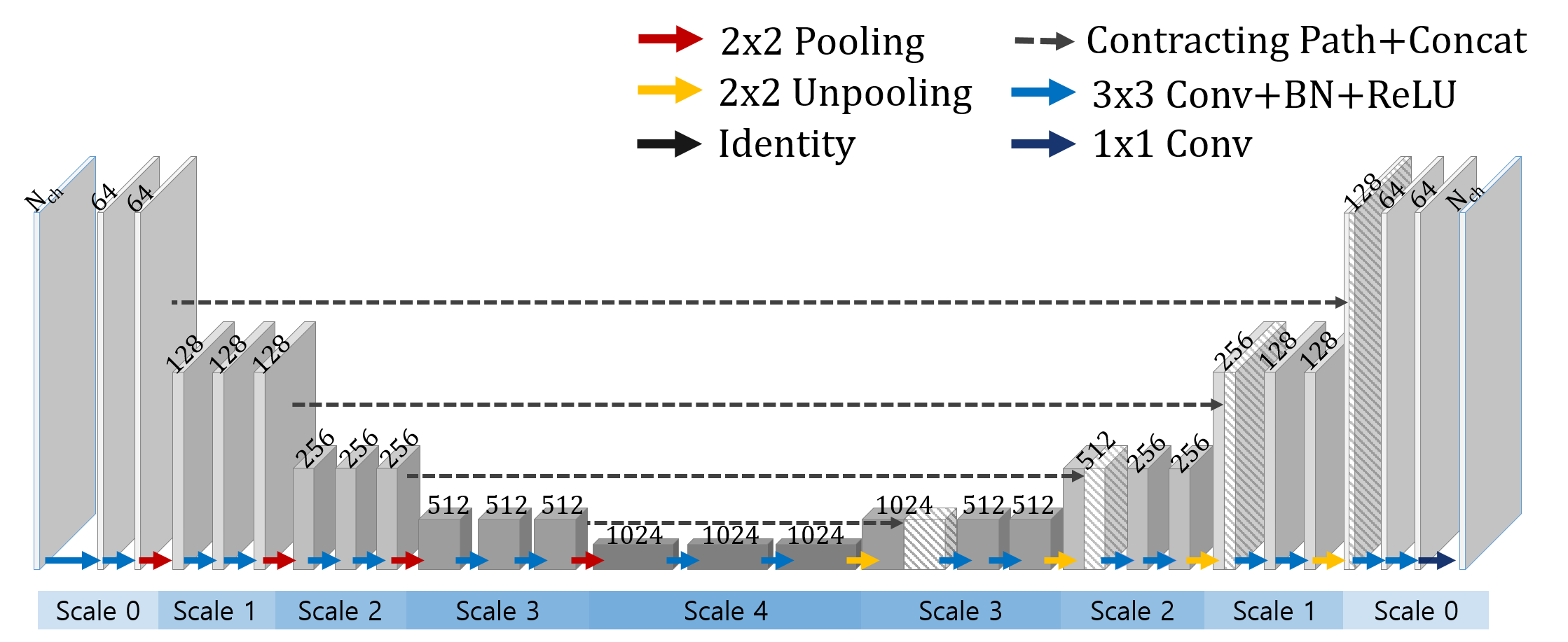}
\caption{The proposed multi-scale  deep network architecture. The input is the aliased image from down-sampled data. The number of channels is displayed at the top of  each block. The network consists of convolution (Conv), batch normalization (BN), ReLU {\color{black}and} contracting path with concatenation layers (dashed arrows). In order to achieve multi-scale representation, the pooling and unpooling  layers are additionally applied. The  scales are displayed under each unit. The output of the network in the former part is concatenated along the channel dimension to the latter part  at the same scale using the contracting path (dashed arrow).}
\label{fig:network}
\end{figure}

Specifically, as shown in {\color{black}Fig. \ref{fig:network}},
the proposed artifact learning network  consists of convolution layer, batch normalization~\cite{ioffe2015batch}, rectified linear unit (ReLU) \cite{krizhevsky2012imagenet}, and contracting path connection with concatenation~\cite{ronneberger2015u}.
More specifically,
 each stage contains  four sequential layers composed of convolution with $3\times3$ kernels, batch normalization and ReLU layers.
 Finally,  the last stage has two sequential layers and the last layer only contains a convolution layer with $1\times1$ kernel.
In the first half of the network, each stage is followed by a max pooling layer, while an unpooling layer
is used in
the later half of the network.
Scale-by-scale contracting paths are used to concatenate the results from the front part of the network to the later part of network. 
The number of channels for each convolution layer is shown in  Fig. \ref{fig:network}.
Note that the number of channels after each pooling layers is doubled.
The resulting multi-scale network has {\color{black}five scales} of representation as shown in Fig.~\ref{fig:network}. The input resolution is 256$\times$256, and the resolution of representation is halved for each change of the scale until it becomes 16$\times$16 in scale 4. Then, with the aid of contracting path with concatenation (the dotted line in Fig.~\ref{fig:network}), each level of information is integrated serially from the low-resolution (scale $k$) into high-resolution representation (original resolution, scale 0). 

The elements of the network are explained in more detail here. 
In the convolutional layer, the weights of convolution play a key role to extract the features of the inputs. 
 Specifically, let $x^l $ denote the $l$-th layer input and $w^l $, $b^l$ represent weights and bias of $l$-th convolution layer, respectively. Then, the convolution layer {\color{black}operates} as follows:
 \begin{eqnarray*}
Conv_{w^l, b^l}(x^l) = w^l * x^l + b^l 
\end{eqnarray*} 
where $*$ is 2D-discrete convolution operation and $w^l$ and $b^l$ have $k_1 \times k_2 \times n_{i} \times n_{o}$ and $n_{o}$ dimension, respectively. Here, $n_{i}$ and $ n_{o}$ are the number for {\color{black}input and output channels}, respectively.
After the convolution layer, we apply batch normalization (BN)~\cite{ioffe2015batch} and the rectified linear unit (ReLU)~\cite{glorot2010understanding}  sequentially to construct a single block of operation, $f^l$:
 \begin{eqnarray}
f^l(x^l; \theta^l) = ReLU( BN_{\gamma^l,\beta^l}( Conv_{w^l,b^l}( x^l ) ) ).
\label{eq:aBlock}
\end{eqnarray} 
where $ReLU(\cdot)=\max(\cdot, 0)$, and $\theta$ consists of convolution filter weights, $(w, b)$, and scaling/shift parameters of batch normalization, $(\gamma, \beta)$. Batch normalization has been widely used in CNN training by incorporating a normalization and scale-shift step before the input of non-linear layer. It stabilizes learning by normalizing the mini-batch of input to have proper mean and variance. It can significantly improve the performance by reducing the problem from poor initialization and helps gradient flow.  ReLU is a necessary non-linear mapping {\color{black}which has an advantage for training the network by reducing the gradient vanishing problem\cite{glorot2010understanding}.}

The skipped connection or contracting path~\cite{he2015deep,ronneberger2015u} is a tool to improve the performance of network by linking or bypassing the results of the previous layer to a downstream layer. 
As shown in Fig.~\ref{fig:network}(a), the dotted line bypasses the result to a {\color{black}latter} layer. There is a difference between skipped connection and contracting path with concatenation as follows:
\begin{eqnarray*}
s( x^{l}, f^{l+k}(x^{l+k})) =  \begin{cases}  
\quad f^{l+k}(x^{l+k})+ x^{l} \quad\quad \quad\quad \quad\quad \text{(skipped connection)} \\
\quad [ f^{l+k}(x^{l+k}), \quad x^{l}] \quad \quad\quad \text{(contracting path with concatenation)}  \end{cases}
\end{eqnarray*} 
More specifically, in the contracting path $s$, the input of $l$-th layer ($x^l$) skips over $k$ layers and 
 is concatenated to the output of $(l+k)$-th layer ($f^{l+k}(x^{l+k})$).
 On the other hand,   $x^l$ is added to the output $f^{l+k}(x^{l+k})$ in  the skipped connection~\cite{he2015deep}. 

The pooling layer reduces the spatial size of the representation to reduce the resolution in the network. It is common 
to use the max pooling layer, and we chose {\color{black} both pooling size and stride as $2\times2$}. Then, the  $(i,j)$-th element of output from the pooling layer operates as follows:
 \begin{eqnarray*}
p(x^l)_{(i,j)} =  \max( P_{(i,j)}^{2\times 2}(x^l) ), \quad \quad i=1,...,n/2, \quad j=1,...,m/2
  \label{eq:pooling}
\end{eqnarray*} 
where $n, m$ are the row and column size of the input representation and $P_{(i,j)}^{2\times 2}$ is the patch extractor of $(i,j)$-th element. The patch extractor is defined as:
\begin{eqnarray*}
P_{(i,j)}^{2\times 2} = (x^l_{(i',j')}, x^l_{(i'+1,j
)}, x^l_{(i',j'+1)}, x^l_{(i'+1,j'+1)} ), 
\quad i'=2i-1,\quad j'=2j-1.
\end{eqnarray*} 
After the pooling layer, the resolution of representation for both x-y dimension is halved. In Fig.~\ref{fig:network}, we denote the change of resolution as a level of scale. 
The unpooling layer is a transpose operation of the pooling layer by upsampling the input with {\color{black}the} rate of {\color{black}two}. 
By utilizing the pooling layer, we can increase the receptive field of the proposed network more efficiently with the same number of convolution. This will be discussed in detail later. Considering that the aliasing {\color{black}artifact} has globally distributed pattern,
the enlarged receptive field from the multi-scale artifact learning is more advantageous {\color{black}for} removing the aliasing artifacts.

\section*{Method}

\subsection*{MR dataset}

We used  brain MR image dataset consisting of total 81 axial brain images of {\color{black}nine} subjects.
The data were acquired in {\color{black}Cartesian} coordinate with a 3T MR scanner with four Rx coils (Siemens, Verio). The following parameters were used for SE and GRE scans: TR 3000-4000ms, TE 4-20ms, slice thickness 5mm, 256$\times$ 256 acquisition matrix, {\color{black}four} coils, FOV 240$\times$ 240, FA 90 degrees. The brain images in the data set have different scales of intensity and maximum values since they were acquired with various scan conditions (GRE/SE, various TE/TR, etc.). Therefore, the data should be normalized for better performance before entering the network. We normalized the data individually to have the same maximum value of 256.
For single-channel experiments, we  selected the first coil data from the four coil data. For parallel imaging experiments, we used all the coil images.
We split the training and test data by  randomly selecting 66  images for training and 15 images for testing.

\subsection*{Down-sampled data generation}

The original k-spaces were retrospectively down-sampled. There are several ways of k-space under-sampling to speed up the 2D MR acquisition. The irregular  and regular {\color{black}samplings} along the phase encoding direction are the examples. 
Recall that our objective is to train the network to learn  the aliasing artifacts.  As will be shown later,  in contrast to the CS-MRI, it is easier to learn aliasing artifact from regular sampling patterns with a few low-frequency
k-space data.
Here, additional low-frequency auto-calibration signal (ACS) lines are necessary to compose the aliasing artifacts mainly from high frequency edge signals rather than low-frequency image repetitions.
So we chose the regular sampling pattern with auto-calibration signal (ACS) lines for down-sampling.
In particular,  this sampling pattern is a common sampling pattern for GRAPPA. Thus, it is not necessary to perform  additional pulse sequence programming. 
Specifically, the k-space data were retrospectively subsampled by a factor of four  with 13 ACS lines (5 percent of total {\color{black}phase encoding lines}) in the k-space center.


In order to make the network more robust,  data augmentation is essential when only a few training data are available. 
In order to produce the augmented MR images, the original full-sampled images are transformed by rotation, shearing and flipping. The transforms were performed on complex domain so that we can acquire the full and down- sampled k-space data of the augmented MR images. By applying the aforementioned transforms,  32 times more training samples were generated for data augmentation.

\subsection*{Magnitude and phase networks}

 We have trained the two artifact networks: one for the magnitude and the other for {\color{black}the} phase.
More specifically, by applying the inverse Fourier transform, we first 
generated the aliased images as inputs to the network. 
Then, for the artifact network to reconstruct  magnitude images, the inputs  of the network were  the magnitude
of distorted MR images and  labels were the magnitude of the aliasing artifact-only images as shown in {\color{black}Fig.~\ref{fig:concept} (c)}. The artifact network for the phase reconstruction was similarly trained. The inputs and labels are {\color{black}the} phases of the distorted images and the aliasing artifact-only images, respectively.
Both networks have the same  structure. 

\begin{figure}[!b] 	
\center{\includegraphics[width=15 cm]
{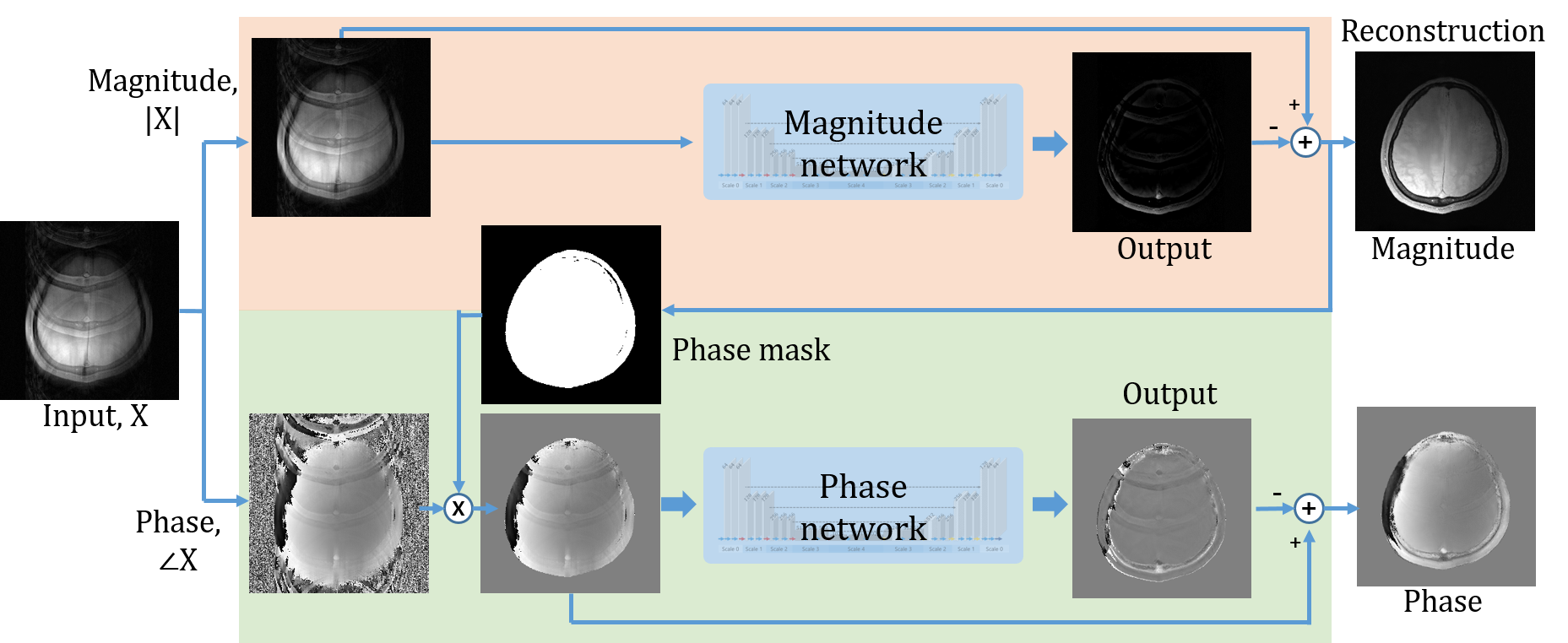}}
\caption{ Reconstruction flow for magnitude and phase images. First, the input, $X$, is converted to the magnitude and the phase images. 
From the magnitude network, the artifact  is estimated in the magnitude image and subtracted from the input image to obtain the final magnitude image (orange block).
Then, we get a phase mask to remove the effect of random phase fluctuations outside the brain. In the phase network (green block),  the phase domain artifact is estimated  within the phase mask, and the final phase reconstruction is performed by subtracting the artifact from the input phase image.
}
\label{fig:recon_flow}
\end{figure}

However, due to the property of phase image, there is an additional step for the phase reconstruction network. 
As shown in  the phase image of Fig.~\ref{fig:recon_flow} (bottom row), the region within the brain has smooth structures and the values of pixels vary slowly from $-\pi$ to $+\pi$. 
{\color{black} While the area outside of the brain has
approximately zero in the magnitude images, they have large fluctuation in the phase images because these phases have random-like values of $-\pi$ to $+\pi$.} These random phases outside of the brain region
 make the network train more difficult.
 

To improve the performance of the network for phase reconstruction, we used the phase masking to remove the effect of that random phases outside of the brain. Specifically, we first trained the magnitude network to get the reconstructed magnitude images. Then the phase masks were obtained from the reconstructed magnitude images using a simple thresholding. Using the phase mask, we can remove the effects of random-phases in the outside of the brain by zeroing {\color{black}out} the outside of the ROI in both the input and the artifact phase images. Then, the phase network is trained by assigning the artifact phase data within phase mask as labels.

After training the two networks, the reconstruction flow follows {\color{black}the} same steps with {\color{black}the} training process as  in Fig.~\ref{fig:recon_flow}.
Since the generation of phase masks should be prior to the phase reconstruction, the magnitude of the MR images is first reconstructed  by the magnitude network (Fig.~\ref{fig:recon_flow} top row).
Then, the estimated aliasing artifact are subtracted from the  distorted input images to generate the final reconstructed images. After the reconstruction of the magnitude images, the phase mask is generated by comparing the pixels of the reconstructed magnitude image to a threshold value. The outside of the aliased brain image is erased with the phase mask to remove the random phases. Then, the artifact of phase image is reconstructed by the proposed phase network and  we could complete the final phase reconstruction by subtracting the artifact  from the input phase image (Fig.~\ref{fig:recon_flow} bottom row).

\subsection*{Implementation details}

The network was implemented with the MatConvNet toolbox (ver.20)~\cite{vedaldi2015matconvnet} in the MATLAB 2015a environment (Mathworks, Natick). We used a GTX 1080 graphic processor and i7-4770 CPU (3.40GHz). 
The weights of the convolutional layers were initialized by Gaussian random distribution with Xavier method\cite{glorot2010understanding} to obtain a correct scale. This has helped us  avoid the signal exploiding or vanishing in the initial phase of learning. The stochastic gradient descent (SGD) method with the momentum was used to train the weights of the network and minimize the loss function. The learning rate was reduced logarithmically from $10^{-2}$ to $10^{-3}$ per epoch. The size of the mini-batch was set to {\color{black}three},  which is the maximum number for the given hardware specification.
It took about 13 hours for training {\color{black}the magnitude network and 9 hours for training the phase network.}

We took the square root of sum of squares (SSOS) on output magnitude images for final reconstruction,  and the SSOS of the magnitude images of full k-spaces data were used as the ground-truth. For the phase images, the phase images from full k-space data were similarly used as the ground-truth.  The reconstruction performance {\color{black}was} measured by the {\color{black}normalized} mean square error (NMSE).

\subsection*{Comparative studies}

To verify the performance of the network, we {\color{black}used} the ALOHA~\cite{jin2016general} reconstruction as the state-of-the-art CS algorithm for both single- and {\color{black}multi-channel} reconstruction.
We also compared the reconstruction results for multi-channel {\color{black}dataset} with those of  GRAPPA~\cite{griswold2002generalized}.

We have also compared the performance of proposed network  with different types of deep networks. 
Specifically, we compared the three learning architectures: (1) image learning with a multi-scale network, (2) artifact learning with a single-scale network and (3) the proposed artifact learning with a multi-scale network (Fig.~\ref{fig:network}(a)) to confirm the importance of artifact learning in a multi-scale manner.
 The main difference in the network architecture between single  and multi-scale is the use of pooling and unpooling. 
In the multi-scale network, both x-y resolutions of input are halved and the number of channels is doubled after each pooling layer (red arrow in Fig.~\ref{fig:network} (a)).
 In the single-scale learning, there were no pooling and unpooling layer, so the same image resolution with the filter depth of  64 channels were used for all  layers.
  The image learning with a multi-scale network could be implemented by simply changing the labels from the artifact images to the original images. 

\section*{Results}\label{sec:result}

\begin{figure}[!b] 	
\centering
\includegraphics[width=16cm]{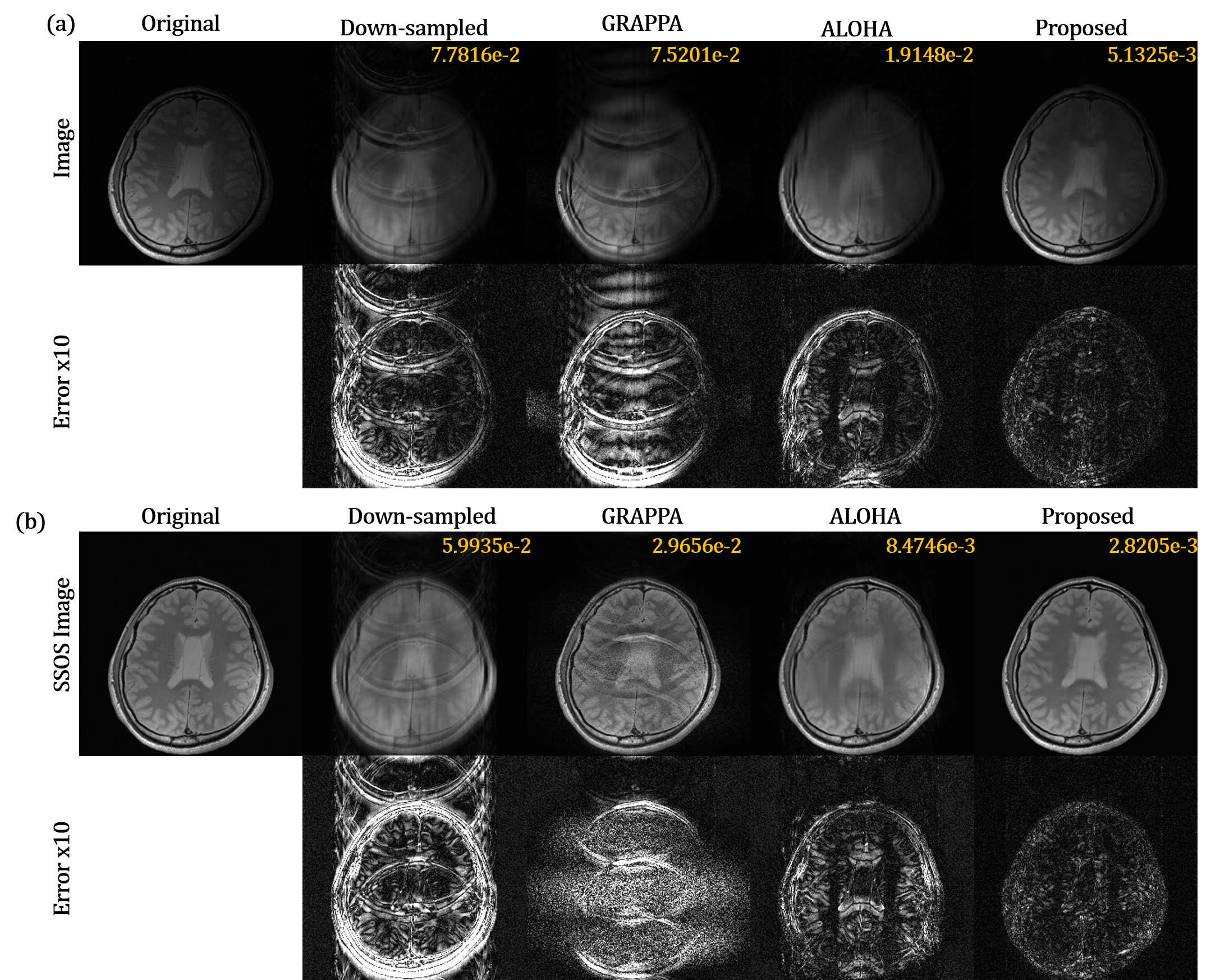}
\caption{ {\color{black}Reconstruction result of the magnitude images for (a) single-channel and (b) multi-channel MR data. The original images are displayed on the first column. The square root of sum of squares (SSOS) are used for multi-channel MR images. The down-sampled rate is $\times$4 with 5 percents of ACS lines. The reconstruction results and its error images are shown in the first and second rows, respectively. The error is $\times$10 amplified for better visualization and NMSE values are displayed on the top of each reconstructed image.}
}

\label{fig:res_mag}
\end{figure}

The performance of the proposed network was first compared with that of ALOHA and GRAPPA. 
The magnitude reconstruction results are displayed in Fig.~\ref{fig:res_mag}. 

In the single channel experiment (Fig.~\ref{fig:res_mag}(a)), there was a significant amount of aliasing artifacts from the zero-filled reconstruction. The difference images {\color{black}show} these aliasing artifacts more clearly.
Since the k-space is uniformly down-sampled, the coherent aliasing artifact  appeared.
However, due to the additional ACS lines, the coherent aliasing artifact are mainly for the edge images.
The result of GRAPPA shows a small improvement, but the most important aliasing artifacts have remained because we have only used one coil image.
{\color{black}Furthermore}, for compressed sensing algorithms, it  was difficult to  remove the aliasing artifacts clearly since compressed sensing is designed for incoherent sampling and artifacts. Most of the existing CS algorithms failed,  but only ALOHA was somewhat successful with some remaining aliasing artifact.
The reconstruction image of ALOHA is better than the zero-filled image and the reconstruction image by GRAPPA visually and quantitatively.
{\color{black}However, the result was still blurry and the aliasing artifacts were remained in the reconstructed image.
In contrast}, the proposed artifact learning algorithm clearly showed accurate reconstruction by removing the coherent aliasing artifacts. As shown in the error image of the proposed method {\color{black}(Fig.~\ref{fig:res_mag})}, the aliasing artifacts are effectively reduced and the NMSE value is minimal compared to the reconstruction results of the above-mentioned methods.

For parallel imaging experiments with four channel data (Fig.~\ref{fig:res_mag} (b)), 
the zero-filled reconstruction images have severe aliasing artifacts and blurred details, but all multi-channel reconstruction showed improvements compared to the single channel experiments.
The GRAPPA reconstruction shows a better result compared to the single channel reconstruction,  but still has many reconstruction errors. As shown in the error image, the reconstruction of GRAPPA still shows remaining aliasings and the noise-like high frequency errors.
ALOHA reconstruction {\color{black}was} able to remove most of the aliasing artifacts, but the results were not perfect due to the coherent sampling.
However, the proposed method provided a great reconstruction results as seen in the Fig.~\ref{fig:res_mag} (b).

\begin{figure}[!b] 		
\centering
\includegraphics[width=16cm]{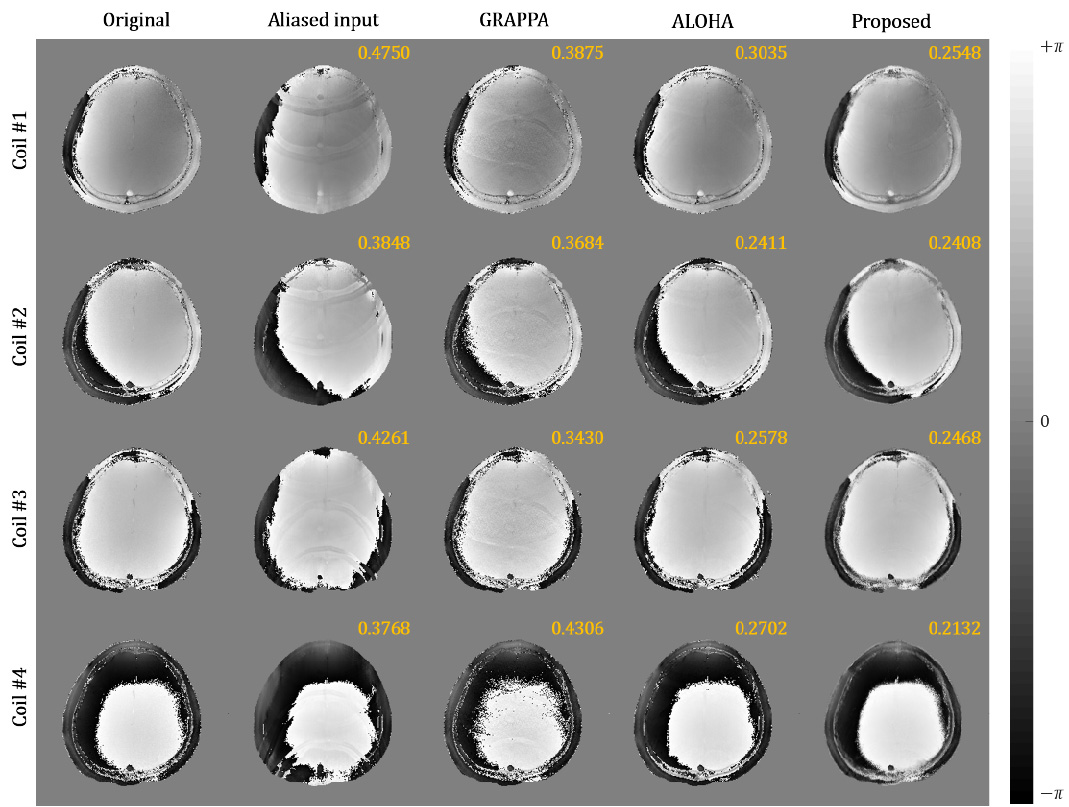}
\caption{ Reconstruction result of phase images. The original images of each channel are displayed in the first column. The reconstruction images of multi-channel MR data are displayed. GRAPPA and ALOHA used multi-channel information for their reconstruction.  The down-sampled rate is $\times4$ with 5 percent of the ACS lines. The reconstruction results of GRAPPA, ALOHA and proposed network are shown in the third, fourth and fifth column. The NMSE values are displayed at the top of each image.
}
\label{fig:res_ang}
\end{figure}

The phase reconstruction results from $\times$4 acceleration are shown in Fig.~\ref{fig:res_ang}. The phase images of the zero-filled reconstruction show strong aliasing artifacts on each coil image. The phase images are usually smooth and they could have discontinuity from $-\pi$ to $+\pi$ due to the phase wrapping.
Although GRAPPA utilizes the multi-coil data, GRAPPA shows poor reconstruction results. The aliasing artifacts are retained and the high frequency error has been enhanced  in the vicinity of the discontinuous regions. And it results {\color{black}in} large NMSE values for each coil image. The phase reconstructions of ALOHA and the proposed method have little aliasing artifacts for all coil images. Compared to ALOHA, the proposed method resulted in minimal errors for each coil image. Even when there is a large signal jump due to the phase wrapping, the proposed network effectively {\color{black}removed} the aliasing artifacts.

Compared to GRAPPA and ALOHA, which are the representitive algorithms in parallel imaging and CS, 
 the proposed method shows better results for both magnitude and phase reconstruction. The reconstruction images using GRAPPA have heavy aliasing artifacts and the enhancement of high frequency  errors (See Fig.~\ref{fig:res_mag} and~\ref{fig:res_ang}). 
This type of imperfect GRAPPA reconstruction usually occurs when the GRAPPA kernel is incorrectly estimated due to the insufficient number of coils and ACS lines.
Although ALOHA  was developed to reconsturct the k-space from irregular sampling pattern, ALOHA somehow reconstructs the images from uniform subsampled k-space data, but they still have strong aliasing artifacts.

\begin{figure}[!b] 	
\center{\includegraphics[width=16 cm]{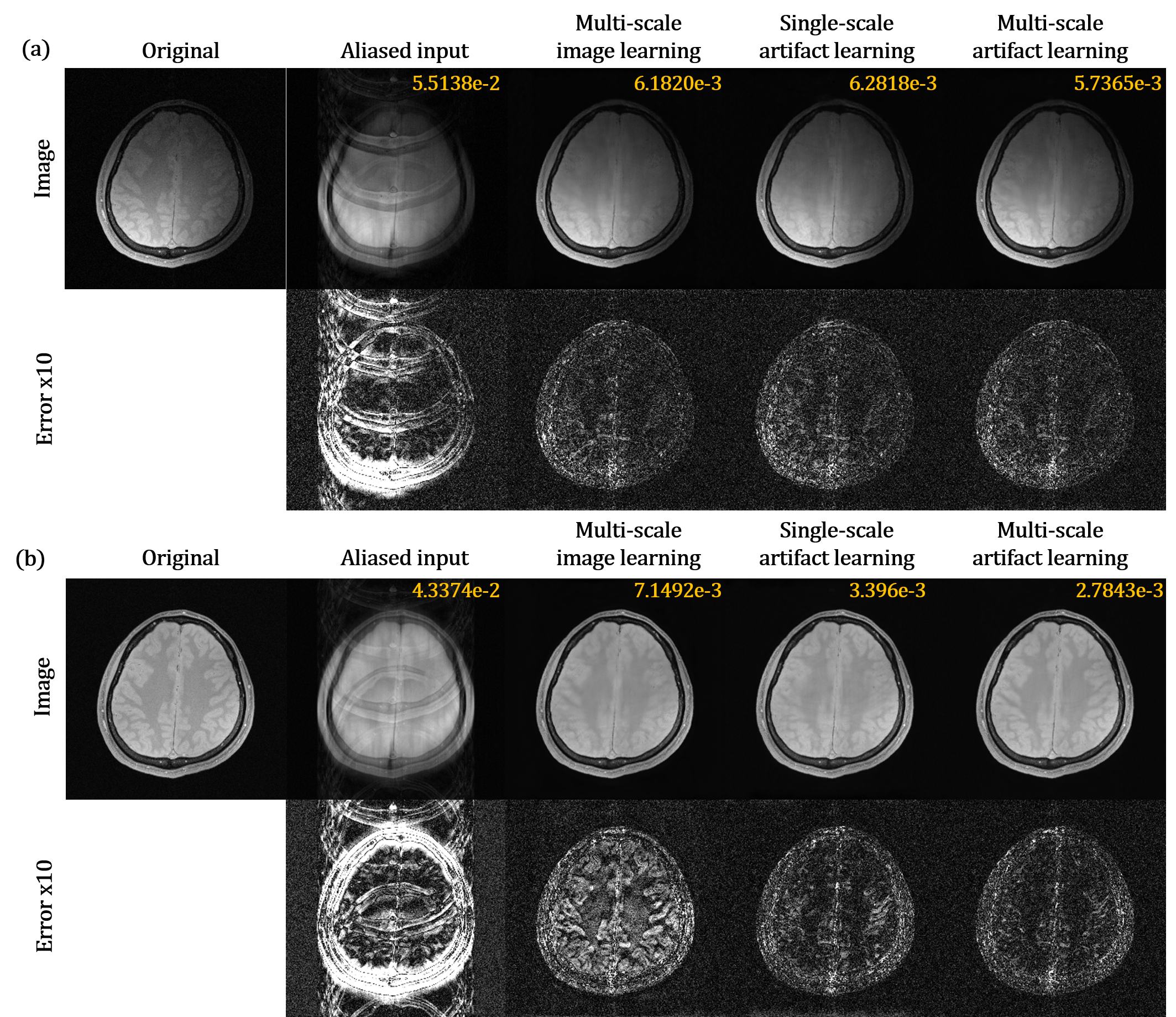}}
\caption{Reconstruction result of magnitude images from (a) single-channel and (b) multi-channel data according to the proposed method. The original images are displayed in the first column. The down sampling rate is 4 with 5 percent of the ACS lines. In the third, fourth and fifth columns the reconstruction results of multi-scale image learning, single-scale artifact learning and multiscale artifact learning (the proposed network) are shown. The NMSE values are displayed at the top of each image.
}
\label{fig:res_net}
\end{figure}

Fig.~\ref{fig:res_net} shows the reconstruction results of three different networks: (1) multi-scale image learning, (2) single-scale artifact learning and (3) the proposed  multi-scale artifact learning.
These three networks {\color{black}showed} much improved reconstruction results compared to GRAPPA and ALOHA {\color{black}in} both single- and multi-channel data. In the single channel results, the proposed network showed  a minimum NMSE value. Similarly, the proposed network showed a significant improvement in the multi-channel result (Fig.~\ref{fig:res_net} (b)). The multi-scale image learning removes the aliasing artifacts, but images are too blurry, causing the large errors on the reconstruction result. Single-scale artifact learning showed much more improved result, which is better than that of the multi-scale image learning. However, the performance of the proposed artifact learning was the best.

We also compared the networks for image learning and artifact learning with the convergence plots for the test {\color{black}dataset} (Fig.~\ref{fig:error_graph}). 
In the magnitude reconstruction for single channel (Fig.~\ref{fig:error_graph} (a)), the two deep networks showed better performance than ALOHA (yellow dashed) and GRAPPA (black dashed). Between the deep networks, the NMSE of the proposed network (red) converged with a minimal error compared to that of the image learning (blue).  We {\color{black}found} similar results for the magnitude reconstruction of multi-channel data (Fig.~\ref{fig:error_graph} (b)). Here, both deep networks {\color{black}showed} better performance compared to the GRAPPA and ALOHA. And the proposed muiti-scale artifact learning showed the best {\color{black}reconstruction performance}.
Compared with the image learning, the proposed artifact learning showed better results for both single and multi-channel reconstruction as shown in Fig.~\ref{fig:error_graph} (a) and (b). 
This strongly suggests that it is better for the network to learn the artifact pattern itself than to learn the original image. 

\begin{figure}[!b] 	
\centering
\includegraphics[width=17cm]{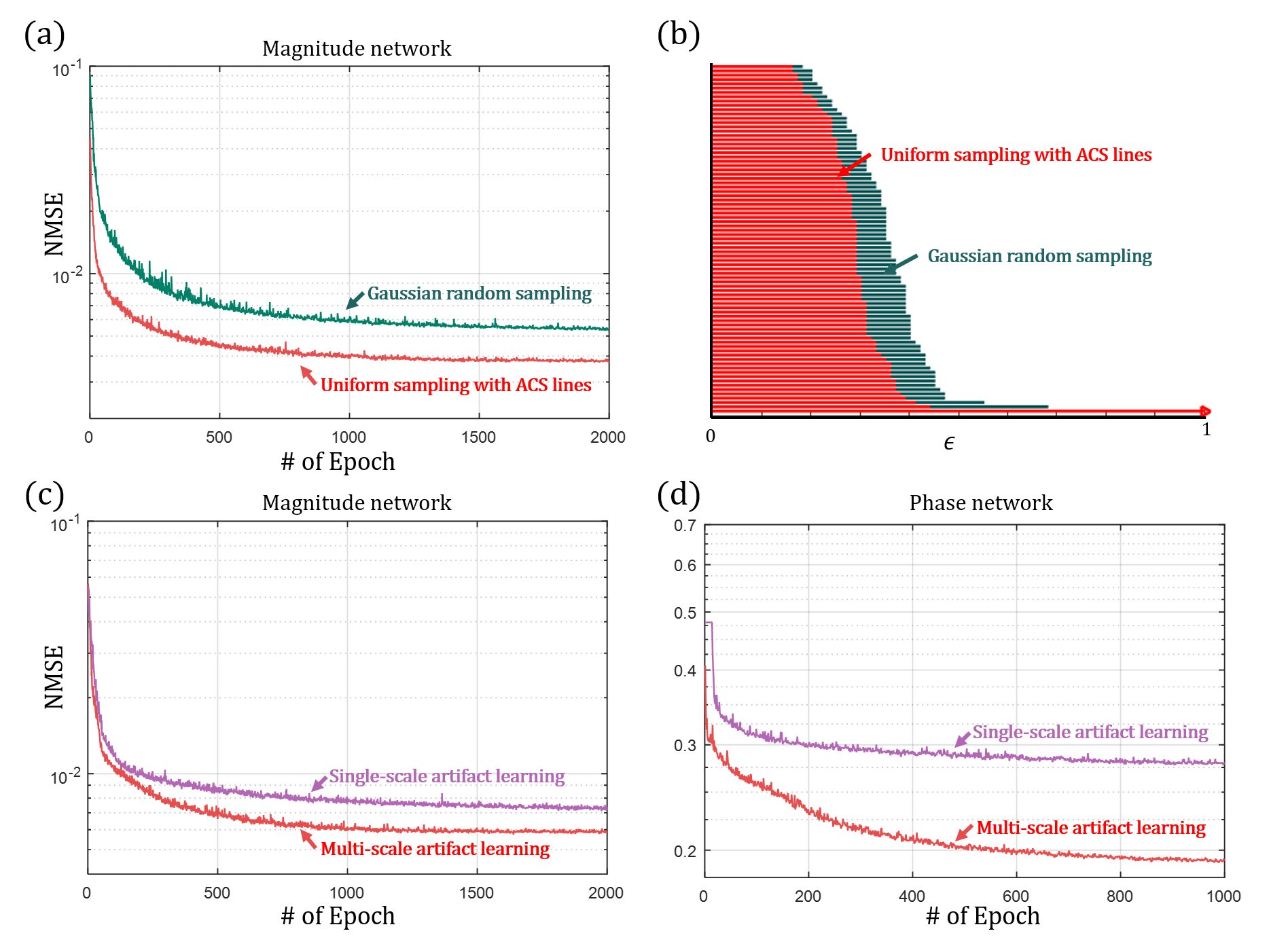}

\caption{ (a) The NMSE plot of proposed network and (b) the zero-dimensional barcode fosr multi-channel MR
data using Gaussian random sampling (green) and uniform sampling with ACS
lines (red). The NMSE plots of single-scale (purple) and multi-scale (red) artifact learning are displayed  for (c)
magnitude and (d) phase image reconstruction using single-channel data. The NMSE values were calculated
from the test data set.
}
\label{fig:error_graph2}
\end{figure}

In Fig.~\ref{fig:error_graph2}(a), 
we compared the two sampling patterns:  uniform sampling with ACS lines and  Gaussian random sampling. As shown in Fig.~\ref{fig:error_graph2} (a), the artifact learning using random sampling has faster convergence and some improvement compared to the original image learning. However, the use of uniform sampling pattern with ACS in artifact learning shows improvement compared to the others.

In Fig.~\ref{fig:error_graph2} (c) and (d), the comparison of single-scale artifact learning and multi-scale artifact learning are also given. For both reconstructions of magnitude and phase {\color{black}images}, the proposed multi-scale network shows much better reconstruction performance. In particular, for the phase reconstruction,  the reconstruction performance of the multi-scale network is much more improved compared to the single-scale one.

%

The reconstruction time of GRAPPA was about 30 seconds for multi-channel data and about 5 seconds for single-channel data under the aforementioned hardware setting.
The reconstruction time for ALOHA was about 10 min for four channel data and about 2 min for single channel data. 
The proposed network required less than 41 ms for a multi-channel image and about 30 ms for a single-channel image.
In the case of phase reconstruction, it takes about 61 ms for each coil since it is necessary to run the amplitude network first to obtain the phase mask.
Since the reconstructions of the individual coil images can be computed in parallel, the total reconstruction time was about 61 ms. Even when 
all coil images are serially reconstructed, the total time for the four phase images was less than 250 ms, which is much shorter than the reconstruction time of the existing CS reconstruction algorithms.

\section*{Discussion}
%
%
%
%
%
%

\subsection*{Persistent homology analysis of data manifold}

To support our claim that the simpler data manifold is better for a deep learning,
we analyzed the topology of the data manifold using  persistent homology.  A persistent homology analysis between the original image data and the aliasing artifact
data was performed on the following datasets: (1) magnitude images of single channel data, (2) magnitude images of multi-channel data, and (3) phase images of single-channel data.

\begin{figure}[!b] 	
\centering
\includegraphics[width=16cm]{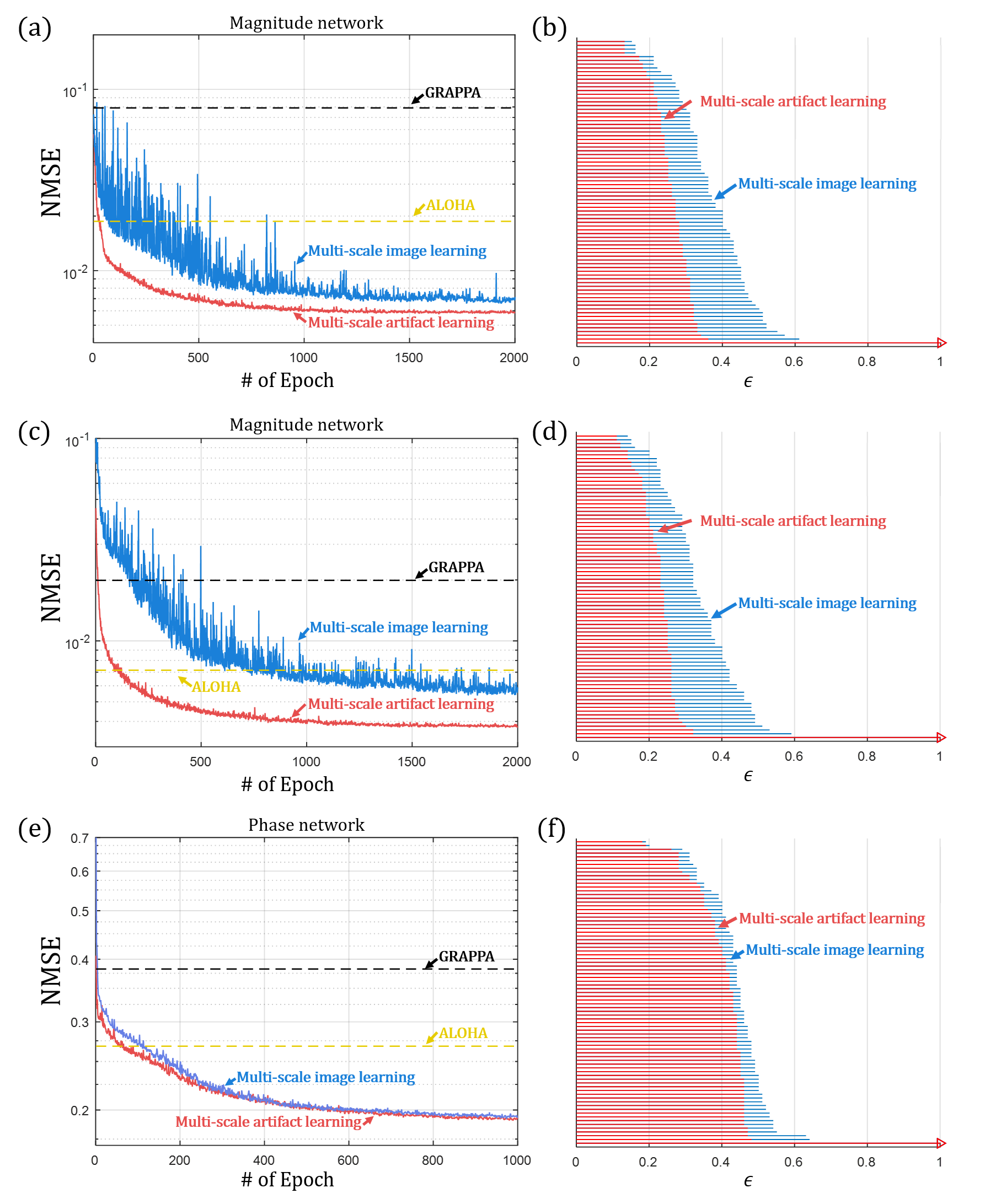}

\caption{ The NMSE plot for (a) magnitude reconstruction from single-channel data, (c) magnitude reconstruction from multi-channel data, and (e) phase reconstruction from single-channel data.  The zero-dimensional barcodes of the original image and the artifact image for each data are displayed in (b), (d) and (f).  The NMSE values were calculated from the test data set.  To compared the reconstruction performance, the NMSE values of GRAPPA (black dashed) and ALOHA (yellow dashed) are displayed together.
}
\label{fig:error_graph}
\end{figure}

In  Fig.~\ref{fig:error_graph} (b), the zero-dimensional barcodes of the artifact data (red) fell faster than the original image data (blue).
This means that the aliasing artifact has a simpler manifold. 
 The simpler manifold resulted in a better performance as shown in the left graph of Fig.~\ref{fig:error_graph} (a). The  errors (red graph) for artifact learning decreased significantly faster than the errors of the image learning (blue graph).
In the case of image learning,  many fluctuations were shown during the initial phase of learning, and it converged slowly.
These large fluctuations became smaller in the final phase of the learning, but the resulting errors were still larger compared to the other convergence plots of artifact learnings.

A similar correspondence between the convergence plot and the persistent homology analysis was found in the multi-channel data {\color{black}(Fig.~\ref{fig:error_graph} (d))}.
The zeroth barcode of multi-channel data indicates that
 the multi-channel image data is more complex than the artifact data.
 As predicted by the persistent homology analysis,  artifact learning converged quickly to a smaller value than the image learning (Fig.~\ref{fig:error_graph} (c)).

The results from the phase image data strongly support a persistent homology analysis as a suitable tool for deep network design.
As can see in the barcode graph of  Fig.~\ref{fig:error_graph} (f), the difference between the barcodes of the artifact and the image data is very small.
This suggests that their topological complexity is similar.
This prediction agreed with the convergence plot in Fig.~\ref{fig:error_graph} (e), which shows that their NMSE convergence plots are similar to each other although there is a slight improvement in the artifact learning.

\subsection*{Uniform versus Random sampling artifact learning}

By using the same network structure, we compared the reconstruction performance {\color{black}between} the use of {\color{black}unfiromly} down-sampled data and Gaussian random down-sampled data.
Previously, using a single coil data, we briefly showed that the artifact from the Gaussian sample has a more complex topological diversity than that of the uniform downsampling with ACS lines while it has a simpler topological manifold than that of the original images (Fig.~\ref{fig:topology} (c)). 
In Fig.~\ref{fig:error_graph2} (b), we have also compared the complexity of the image and artifact manifolds for multi-channel data. 
The uniform sampling with ACS lines again brought  a simpler manifold.
This is consistent with the reconstruction results showing that the artifact-learning with
 the uniformly down-sampled data with the ACS lines has   produced  much better performance than other data sets,
 as shown in Fig.~\ref{fig:error_graph2} (a).
It is believed that the regular repetition of the artifact in the uniform down-sampled data can help  train CNN more effectively.

\subsection*{Multi-scale versus single{\color{black}-}scale artifact learning}

\begin{figure}[!b] 	
\centering
\includegraphics[width=16cm]{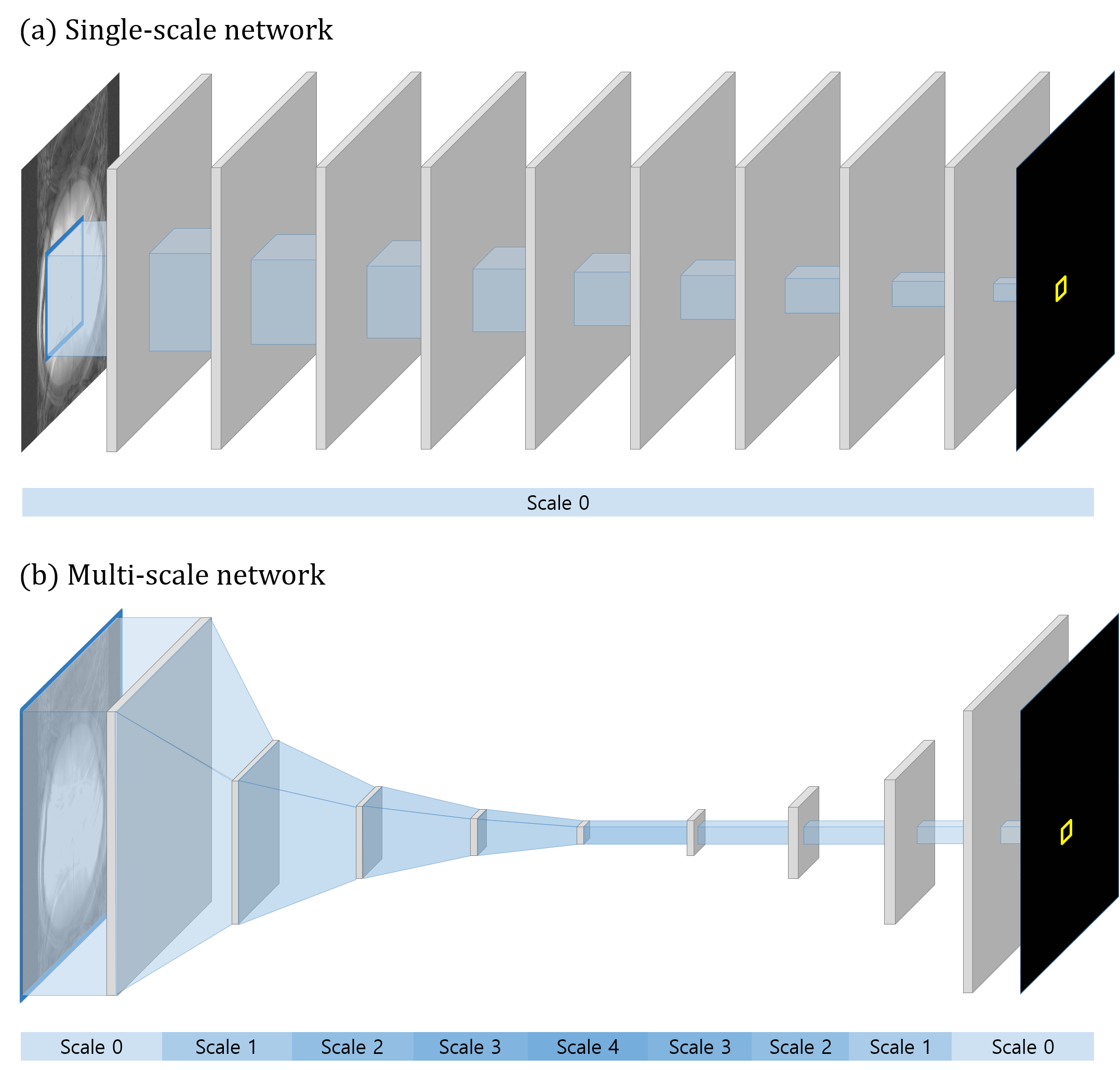}

\caption{Visualization of the receptive fields for (a) single-scale and (b) multi-scale networks.
Each blue square area of the input image is used to reconstruct the yellow square of the output. Compared to the linear increases in the receptive field size from successive convolution layers in the single-scle network, the receptive field increases exponentially through the pooling layers in the case of a multi-scale network. Accordingly, the receptive field of the multi-scale network completely covers the input as opposed to the single-scale network.
}
\label{fig:rField}
\end{figure}

To illustrate the importance of the large receptive field, we compared the multi-scale and the single-scale artifact learning.
Fig. \ref{fig:rField} compares the variation of depth-wise receptive field for a simplified form of the single-scale network (a) and the proposed multi-scale network (b).
Both the single-scale and the multi-scale network consist of 18 layers of 3$\times$3 convolution filters. Accordingly, the size of the final receptive field in single-scale network was 37$\times$37, while  the receptive field of multi-scale network fully covered the 256$\times$256 size of inputs.
The receptive field of the single-scale network increases linearly through the series of convolution layers. 
On the other hand, by using the pooling layers, the receptive field in the multi-scale network increases 
exponentially. 
As a result,  the receptive field of the proposed network with the same number of convolution layers completely covers the entire input as opposed to the single-scale network.
Specifically, each blue square of the input  in Fig.~\ref{fig:rField} is used to reconstruct the yellow square of the output,
showing that  the multi-scale network could more effectively learn the globally distributed artifact patterns.

This is confirmed in Fig.~\ref{fig:error_graph2}(c) and (d), where multi-scale approaches gives the minimal errors. 
The advantage of multi-scale reconstruction was much clearer in  phase reconstruction (Fig.~\ref{fig:error_graph2} (d)). 
The single-scale learning of the phase showed a poor reconstruction result, while  multi-scale learnings showed smaller error in the phase reconstruction.
 Since the phase images are smooth and slowly changing, the bulk of the artifact comes from a globally distributed pattern instead of local one.  This property of phase images is suitable for the application of the multi-scale network, which leads to a great performance improvement in the phase reconstruction (Fig.~\ref{fig:error_graph2} (d)).

This agrees with our earlier works on the x-ray CT application of deep learning \cite{han2016deep, kang2016deep}.  Multi-scale network is more effective in eliminating globally distributed streaking artifacts from sparse projection views \cite{han2016deep}, while the single-scale network is better for removing locally distributed noise from low-dose CT \cite{kang2016deep}.
Because the aliasing artifacts in compressed sensing MRI is more globally distributed,  the multi-scale network is more effective.

\section*{Conclusion}\label{sec:conclusion}

This paper proposed  a deep artifact learning network for the reconstruction of MR images from accelerated MR acquisition. Based on the observation that the aliasing artifacts from {\color{black}uniformly} subsampled k-space data with additional AC lines at low frequency have a simpler data manifold compared to {\color{black}the} other sampling patterns and the artifact-free images,  our network was designed to learn the artifact patterns instead of artifact-free images.

Our experimental results confirmed that the performance of deep  network depends on the topological complexity of the label data manifold. In order to cope with the globally distributed artifact patterns,  the proposed network  also utilized the multi-scale network structure called U-net  having a {\color{black}large} receptive field. We have also confirmed that the multi-scale approach with U-net architecture exhibited a better reconstruction performance than the single-scale network.
In particular, the advantage of the multi-stage network was more pronounced in phase reconstruction, where most of the artifacts are global aliasing patterns rather than localized errors.

Although the training of the network takes a long time, the training could  only  be carried out  once by the manufacturer, so in real application scenario,  network runs very fast and the reconstruction can be performed quickly at each scanner. The very short reconstruction time was one of the great advantages compared to the CS-based iteration reconstruction methods. 

In summary,  the proposed method operated on not only with multi-channel data but also with single-channel data. Even with strong coherent aliasing artifacts, the proposed artifact learning network has successfully learned the aliasing artifacts while the existing parallel and CS reconstruction method have not been able to remove the aliasing artifacts. 
The significant advantages of both computational time and the reconstruction quality suggest that the proposed deep artifact learning is a promising research direction for accelerated MRI with great potential impact.

\section*{Acknowledgement}

This study was supported by Korea Science and Engineering Foundation under Grant NRF-2016R1A2B3008104.

%


\begin{thebibliography}{10}

\bibitem{pruessmann1999sense}
Pruessmann~KP, Weiger~M, Scheidegger~MB, Boesiger~P et~al.
\newblock {SENSE: sensitivity encoding for fast MRI}.
\newblock Magnetic resonance in medicine 1999; 42:952--962.

\bibitem{griswold2002generalized}
Griswold~MA, Jakob~PM, Heidemann~RM, Nittka~M, Jellus~V, Wang~J, Kiefer~B,
  Haase~A.
\newblock Generalized autocalibrating partially parallel acquisitions
  ({GRAPPA}).
\newblock Magnetic resonance in medicine 2002; 47:1202--1210.

\bibitem{donoho2006compressed}
Donoho~DL.
\newblock Compressed sensing.
\newblock IEEE Transactions on information theory 2006; 52:1289--1306.

\bibitem{lustig2008compressed}
Lustig~M, Donoho~DL, Santos~JM, Pauly~JM.
\newblock Compressed sensing {MRI}.
\newblock IEEE signal processing magazine 2008; 25:72--82.

\bibitem{jin2016general}
Jin~KH, Lee~D, Ye~JC.
\newblock A general framework for compressed sensing and parallel {MRI} using
  annihilating filter based low-rank hankel matrix.
\newblock IEEE Trans. on Computational Imaging 2016; 2:480--495.

\bibitem{lee2016acceleration}
Lee~D, Jin~KH, Kim~EY, Park~SH, Ye~JC.
\newblock {Acceleration of MR parameter mapping using annihilating filter-based
  low rank hankel matrix (ALOHA)}.
\newblock Magnetic resonance in medicine 2016; 76:1848?1868.

\bibitem{lee2016reference}
Lee~J, Jin~KH, Ye~JC.
\newblock {Reference-free single-pass EPI Nyquist ghost correction using
  annihilating filter-based low rank Hankel matrix (ALOHA)}.
\newblock Magnetic resonance in medicine 2016; 76:1775?1789.

\bibitem{ye2016compressive}
Ye~JC, Kim~JM, Jin~KH, Lee~K.
\newblock Compressive sampling using annihilating filter-based low-rank
  interpolation.
\newblock IEEE Transactions on Information Theory 2017; 63:777--801.

\bibitem{krizhevsky2012imagenet}
Krizhevsky~A, Sutskever~I, Hinton~GE.
\newblock Imagenet classification with deep convolutional neural networks.
\newblock { In:} Advances in Neural Information Processing Systems, { In:}
  Advances in Neural Information Processing Systems, 2012.  pp. 1097--1105.

\bibitem{ronneberger2015u}
Ronneberger~O, Fischer~P, Brox~T.
\newblock U-net: Convolutional networks for biomedical image segmentation.
\newblock { In:} International Conference on Medical Image Computing and
  Computer-Assisted Intervention, { In:} International Conference on Medical
  Image Computing and Computer-Assisted Intervention, 2015.  pp. 234--241.

\bibitem{mao2016image}
Mao~XJ, Shen~C, Yang~YB.
\newblock Image denoising using very deep fully convolutional encoder-decoder
  networks with symmetric skip connections.
\newblock arXiv preprint 2016; .

\bibitem{zhang2016beyond}
Zhang~K, Zuo~W, Chen~Y, Meng~D, Zhang~L.
\newblock Beyond a {Gaussian} denoiser: Residual learning of deep {CNN} for
  image denoising.
\newblock arXiv preprint arXiv:1608.03981 2016; .

\bibitem{dong2014learning}
Dong~C, Loy~CC, He~K, Tang~X.
\newblock Learning a deep convolutional network for image super-resolution.
\newblock { In:} European Conference on Computer Vision, { In:} European
  Conference on Computer Vision, 2014.  pp. 184--199.

\bibitem{vapnik1998statistical}
Vapnik~VN, Vapnik~V, ``Statistical learning theory'', Vol.~1. Wiley New York,
  1998.

\bibitem{telgarsky2016benefits}
Telgarsky~M.
\newblock Benefits of depth in neural networks.
\newblock arXiv preprint arXiv:1602.04485 2016; .

\bibitem{bianchini2014complexity}
Bianchini~M, Scarselli~F.
\newblock On the complexity of neural network classifiers: A comparison between
  shallow and deep architectures.
\newblock IEEE Trans. on Neural Networks and Learning Systems 2014;
  25:1553--1565.

\bibitem{kang2016deep}
Kang~E, Min~J, Ye~JC.
\newblock A deep convolutional neural network using directional wavelets for
  low-dose x-ray ct reconstruction.
\newblock arXiv preprint arXiv:1610.09736 2016; .

\bibitem{jin2016deep}
Jin~KH, McCann~MT, Froustey~E, Unser~M.
\newblock Deep convolutional neural network for inverse problems in imaging.
\newblock arXiv preprint arXiv:1611.03679 2016; .

\bibitem{han2016deep}
Han~Y, Yoo~J, Ye~JC.
\newblock Deep residual learning for compressed sensing {CT} reconstruction via
  persistent homology analysis.
\newblock arXiv preprint arXiv:1611.06391 2016; .

\bibitem{wang2016accelerating}
Wang~S, Su~Z, Ying~L, Peng~X, Zhu~S, Liang~F, Feng~D, Liang~D.
\newblock Accelerating magnetic resonance imaging via deep learning.
\newblock { In:} 2016 IEEE 13th International Symposium on Biomedical Imaging
  (ISBI), { In:} 2016 IEEE 13th International Symposium on Biomedical Imaging
  (ISBI), 2016.  pp. 514--517.

\bibitem{hammernik2016learning}
Hammernik~K, Knoll~F, Sodickson~D, Pock~T.
\newblock Learning a variational model for compressed sensing {MRI}
  reconstruction.
\newblock { In:} Proceedings of the International Society of Magnetic Resonance
  in Medicine (ISMRM), { In:} Proceedings of the International Society of
  Magnetic Resonance in Medicine (ISMRM), 2016.

\bibitem{chen2015learning}
Chen~Y, Yu~W, Pock~T.
\newblock On learning optimized reaction diffusion processes for effective
  image restoration.
\newblock { In:} Proceedings of the IEEE Conference on Computer Vision and
  Pattern Recognition, { In:} Proceedings of the IEEE Conference on Computer
  Vision and Pattern Recognition, 2015.  pp. 5261--5269.

\bibitem{chen2015trainable}
Chen~Y, Pock~T.
\newblock Trainable nonlinear reaction diffusion: A flexible framework for fast
  and effective image restoration.
\newblock arXiv preprint arXiv:1508.02848 2015; .

\bibitem{noh2015learning}
Noh~H, Hong~S, Han~B.
\newblock Learning deconvolution network for semantic segmentation.
\newblock { In:} Proceedings of the IEEE International Conference on Computer
  Vision, { In:} Proceedings of the IEEE International Conference on Computer
  Vision, 2015.  pp. 1520--1528.

\bibitem{he2015deep}
He~K, Zhang~X, Ren~S, Sun~J.
\newblock Deep residual learning for image recognition.
\newblock arXiv preprint arXiv:1512.03385 2015; .

\bibitem{lustig2010spirit}
Lustig~M, Pauly~JM.
\newblock {SPIRiT: Iterative self-consistent parallel imaging reconstruction
  from arbitrary k-space}.
\newblock Magnetic resonance in medicine 2010; 64:457--471.

\bibitem{bartlett2002rademacher}
Bartlett~PL, Mendelson~S.
\newblock Rademacher and {G}aussian complexities: {R}isk bounds and structural
  results.
\newblock Journal of Machine Learning Research 2002; 3:463--482.

\bibitem{edelsbrunner2008persistent}
Edelsbrunner~H, Harer~J.
\newblock Persistent homology-a survey.
\newblock Contemporary Mathematics 2008; 453:257--282.

\bibitem{ioffe2015batch}
Ioffe~S, Szegedy~C.
\newblock Batch normalization: Accelerating deep network training by reducing
  internal covariate shift.
\newblock arXiv preprint arXiv:1502.03167 2015; .

\bibitem{glorot2010understanding}
Glorot~X, Bengio~Y.
\newblock Understanding the difficulty of training deep feedforward neural
  networks.
\newblock { In:} Aistats, { In:} Aistats, 2010.  pp. 249--256.

\bibitem{vedaldi2015matconvnet}
Vedaldi~A, Lenc~K.
\newblock Matconvnet: Convolutional neural networks for {MATLAB}.
\newblock { In:} Proceedings of the 23rd ACM international conference on
  Multimedia, { In:} Proceedings of the 23rd ACM international conference on
  Multimedia, 2015.  pp. 689--692.

\end{thebibliography}

\clearpage

\epstopdfsetup{suffix=-eps2pdf}
\section*{List of figures}

\clearpage

\clearpage

\clearpage



\clearpage

\clearpage

\clearpage

\clearpage

\clearpage

\clearpage


\clearpage


\end{document}